# Generalizable field of view extension for diffusion MRI via conditional flow matching

Gaurav Rudravaram*[a], Lianrui Zuo[a], Ema Topolnjak[a], Michael E. Kim[b], Adam M. Saunders[a], Elyssa M. McMaster[a], Paula Diaz Trujillo[c], Ely E. Wesley[d,e,f], Colin D. McKnight[g], Pratik P. Pandharipande[d,e], Mayur B. Patel[h,e], Cayce Strength[e], Brenda T. Pun[d,e], Amelia W Maiga[e,h], Robert S. Dittus[i,f], David A. Bennett[j], Julie A. Schneider[j], Konstantinos Arfanakis[j,l], The HABS-HD Study Team[k], Derek B. Archer[c], Timothy J. Hohman[c], Yelena Bodien[e,h], Daniel C. Moyer[b], Kurt G. Schilling[g], Bennett A. Landman[a,b]

[a]Dept. of Electrical and Computer Engineering, Vanderbilt University, Nashville, TN USA; [b]Dept. of Computer Science, Vanderbilt University, Nashville, TN USA; [c]Dept. of Neurology, Vanderbilt Health, Nashville, TN USA; [d]Division of Allergy, Pulmonary and Critical Care, Vanderbilt Health, Nashville, TN USA; [e]The Critical Illness, Brain Dysfunction and Survivorship (CIBS) Center; [f]Tennessee Valley Veteran's Affairs Geriatric Research Education Clinical Center (GRECC), Nashville, TN; [g]Dept. of Radiology and Radiological Sciences, Vanderbilt Health, Nashville, TN USA; [h]Division of Acute Care Surgery, Vanderbilt Health, Nashville, TN, USA; [i]Division of General Internal Medicine and Public Health, Vanderbilt Health; [j]Rush University Medical Center, Chicago, IL, USA; Nashville, TN, USA; [k]HABS-HD MPIs: Sid E O'Bryant, Kristine Yaffe, Arthur Toga, Robert Rissman, & Leigh Johnson; and the HABS-HD investigators: Meredith Braskie, Kevin King, James R Hall, Melissa Petersen, Raymond Palmer, Robert Barber, Yonggang Shi, Fan Zhang, Rajesh Nandy, Roderick McColl, David Mason, Bradley Christian, Nicole Joe Lee, Badri Vardarajan, Monica Rivera Mindt, Phillips, Stephanie Large, Amrita Cheema, Lisa Barnes, Mark Mapstone, Annie Cohen, Amy Kind, Ozioma Okonkwo, Raul Vintimilla, Zhengyang Zhou, Michael Donohue, Rema Raman, Matthew Borzage, Michelle Mielke, Beau Ances, Ganesh Babulal, Jorge Llibre-Guerra, Carl Hill and Rocky Vig; [l]Department of Biomedical Engineering, Illinois Institute of Technology, Chicago, IL, USA

## ABSTRACT

Incomplete field-of-view (FOV) diffusion MRI (dMRI) acquisitions compromise whole-brain white matter analysis by excluding brain regions from tractography and connectivity studies. While deep generative models for dMRI FOV extension have been proposed, their generalizability remains limited by the acquisition-dependent nature of dMRI, where signal intensity varies with b-value, gradient direction, and scanner protocol, and the inherent 4D complexity of the data, making it difficult to train models that transfer across sites or acquisition schemes without retraining. We propose a flow-matching generative model that extends the dMRI FOV in both the superior and inferior directions by predicting apparent diffusion coefficients (ADC) in six directions, conditioned on the two nearest acquired slices and the full-FOV T1w image to enforce both local diffusion continuity and anatomical consistency. The ADC representation decouples the model from acquisition-specific b-values and gradient directions, providing a middle-ground across protocols. The model was trained on 3,103 subjects (5,105 sessions) across five different datasets spanning diverse acquisition schemes (b = 500 and 1000 s/mm²; 21–192 gradient directions), four single-shell (b = 1000) and one multi-shell (b = 500/1000). Evaluated on 299 subjects across five independent datasets spanning b = 750–1500 s/mm² and 32–90 gradient directions, with single- and multi-shell schemes unseen during training, the proposed method achieves significantly improved bundle Dice score on downstream tractography ($p < 0.01$). To facilitate reproducibility and adoption, we publicly release the trained model weights together with a containerized inference pipeline that performs automatic FOV extension from truncated dMRI and co-registered T1-weighted images [zenodo link will be provided for final submission].

*gaurav.rudravaram@vanderbilt.edu

## 1. INTRODUCTION

Alterations in white matter have been reported across a wide range of neurological and psychiatric conditions, including Alzheimer's disease (AD)[1], traumatic brain injury[2,3], delirium[4], bipolar disorder[5], and age-related cognitive decline[6]. In particular, degradations in white matter integrity have been implicated early during AD and are shown to be associated with progressive cognitive impairment[7]. Although macrostructural alterations, such as regional atrophy and brain volume loss, have been extensively characterized[8], growing evidence suggests that subtle microstructural abnormalities emerge during healthy aging and the earliest stages of neurodegeneration[9]. Characterizing these microstructural alterations and their trajectories across the lifespan[10] is therefore essential for improving our understanding of brain aging, disease progression, and the mechanisms underlying cognitive decline.

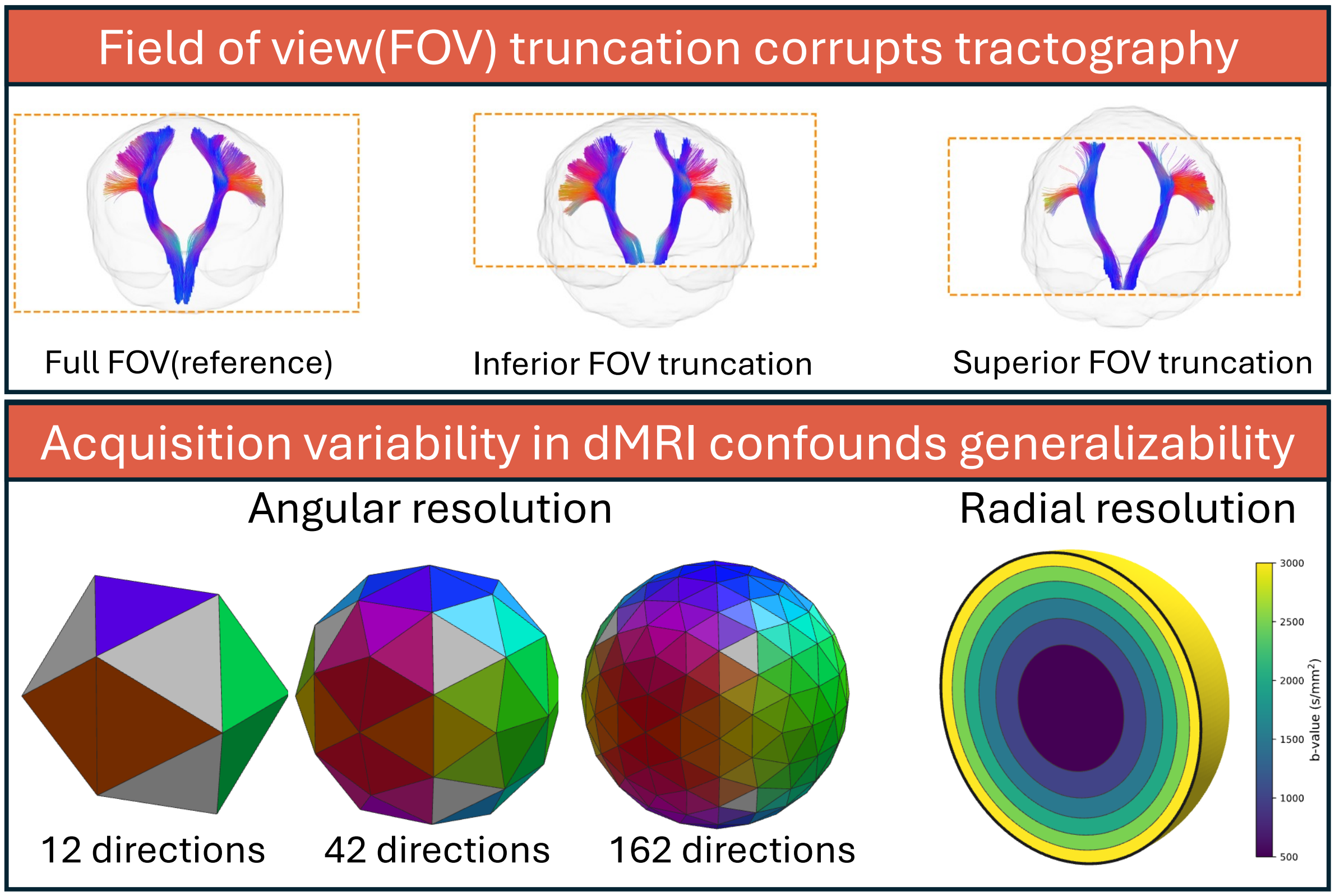

Figure 1: *Top row:* White matter bundles and tracts that span the inferior and superior portions of the brain are impacted due to FOV truncation. Bundles that extend toward the top and bottom of the brain are the most affected. In the full-FOV subject (left), the corticospinal tract (CST) fans out toward the cortex superiorly and reaches the brainstem inferiorly. Under inferior truncation (middle), the CST is cut off before reaching the spine and under superior truncation (right), it fans out less and does not extend to the top of the brain. Missing segments like these can propagate into downstream tract-based and connectivity analyses. *Bottom row:* Representational deep learning for dMRI is complicated by differing acquisitions differ in angular resolution (shown as icosahedral tessellations) and radial resolution (the number and value of sampled b-value shells.

Diffusion magnetic resonance imaging (dMRI) enables non-invasive characterization of white matter microstructure by sensitizing the MR signal to the microscopic displacement of water molecules inside tissue[11]. The diffusion weighting is achieved by applying paired diffusion-sensitizing gradients, causing the measured signal attenuation to depend on the self-diffusion of water molecules[12]. The degree of diffusion weighting is determined by the b-value, which depends on the strength, duration, and timing of the applied gradients[13]. In white matter, water diffusion is anisotropic due to the restrictive effects of axonal membranes and myelin, resulting in signal attenuation that depends on the orientation of the applied diffusion gradient relative to the underlying fiber architecture[14]. Repeating these measurements along many directions probes the orientational structure of white matter, and the diffusion signal in each voxel is naturally described on a solid sphere, with a radial component set by the strength of the diffusion weighting and an angular component set by the direction along which it is applied[15,16]. This orientational information estimated at each voxel can then be propagated throughout the brain using tractography to reconstruct white matter pathways enabling automated segmentation and analysis of anatomically defined fiber bundles with publicly available methods such as TractSeg[17], BundleParc[18] and BrainstemSeg[19].

However, adequate angular and radial sampling requires acquiring multiple diffusion-weighted volumes, so dMRI acquisition times exceed those of a structural T1-weighted (T1w) scans[20]. To remain feasible, diffusion protocols generally adopt coarser spatial resolution and a more restricted field-of-view (FOV) compared to the accompanying T1w scans. The reduced coverage leaves little margin for error in slab placement, and variability in positioning, head size, or subject motion truncates the FOV at the superior or inferior extent of the brain, most often affecting the vertex and cerebellum[21]. Truncation removes the diffusion information required to reconstruct any streamline passing through the affected region, causing bundles that traverse those areas to terminate prematurely or be missed entirely during segmentation (Figure 1). Bundle-wise measures are then computed over incomplete pathways, introducing systematic bias and, in longitudinal or multi-site studies, variability unrelated to the biology of interest. Recovering the missing diffusion signal is essential for preserving tractography and unbiased bundle-wise microstructural analyses, motivating learning-based approaches for FOV extension.

Deep generative models provide a promising route to recovering the missing diffusion signal. Unlike conventional anatomical images, however, diffusion MRI presents a unique representational challenge because acquisition protocols vary substantially across studies. Since the signal is jointly defined by its radial sampling (b-values) and angular sampling (gradient directions), measurements from different protocols differ in dimensionality, in the directions sampled, and in absolute scale. This protocol dependence has motivated acquisition-invariant representations of the diffusion signal. Golkov et al[22]. mapped raw q-space measurements to microstructural parameters, but the mapping remains tied to the sampling scheme it was trained on. Spherical harmonic expansions provide a continuous, fixed-length alternative and have been applied to cross-scanner harmonization[23] and FOD estimation[24], though high-order expansions require dense angular sampling to be well-conditioned. Nath et al.[25] used a SHORE basis to obtain multi-shell b-value generalizability. Yao et al.[26] convert the measured signal to apparent diffusion coefficients and resample onto the vertices of an icosahedral tessellation by inverse-distance interpolation, showing that a single model can then operate across datasets with differing gradient schemes. Collectively, these studies demonstrate that suitable signal representations can substantially reduce dependence on acquisition-specific sampling schemes.

Despite these advances in protocol-invariant signal representations, existing learning-based FOV extension methods for dMRI remain largely acquisition-specific. Chen et al.[27] parcellate tractography under incomplete FOV by augmenting training with synthetically truncated tractograms, allowing streamlines to be correctly labeled despite missing regions. The missing signal is never recovered, so the approach must be re-engineered for each downstream analysis and cannot restore microstructural measures in the affected territory. Gao et al.[21] synthesize missing diffusion-weighted volumes with a conditional pix2pix framework, mapping each volume independently using a co-registered T1w image as anatomical reference and Li et al.[28] extend this with a conditional variational U-Net that additionally conditions on the gradient table. Both are trained at a single diffusion weighting and require retraining for acquisitions with different radial sampling. As a result, existing approaches cannot be readily applied to the heterogeneous acquisition protocols commonly encountered in multi-site studies and clinical practice.

In this work we propose a conditional flow matching model for dMRI FOV extension that is designed to generalize across acquisition protocols. We introduce a protocol-independent six-channel ADC representation obtained by fitting the diffusion tensor to the full acquisition and evaluating it along six canonical icosahedral directions, mapping any acquisition, at any b-value or gradient scheme, onto a common target of fixed dimensionality that is invertible back to the tensor. We then formulate FOV extension as conditional rectified flow, conditioning on the two nearest acquired ADC slices to enforce diffusion continuity and on the co-registered full-FOV T1w slice to supply the anatomy of the target region and extend autoregressively in either the superior or inferior direction without an explicit direction encoding.

## 2. METHODS

### 2.1 Data and preprocessing

To evaluate whether FOV extension generalizes across diverse diffusion MRI acquisition protocols, we curated a training dataset spanning a broad range of angular and radial resolutions. The training set consists of 3,103 subjects across 5,105 imaging sessions from five studies: Health and Aging Brain Study- Health Disparities (HABSHD)[29], Indiana, KBASE[30], ROS/MAP/MARS[31], and ABCD[32]. Ten percent of the training data was held out for model selection, with the split performed at the subject level rather than the session level to prevent data leakage across repeated imaging sessions. Generalization was evaluated on five studies entirely excluded from training, each with a distinct cohort and acquisition: HCP(healthy young adults, high angular and radial sampling), FloridaADRC and WRAP(aging and Alzheimer's disease), VUIIS-MS(multiple sclerosis; IRB 141337), and BRAIN2 (delirium and survivorship in critically ill patients; IRB 192003).These datasets encompass acquisition protocols with angular and radial resolutions that differ from those represented in the training set (Table 1). There was no subject overlap between the training and test sets, ensuring that performance reflects the model's ability to generalize to previously unseen subjects acquired under unfamiliar imaging protocols. Furthermore, the test set was curated to include only a single imaging session per subject.

All diffusion MRI (dMRI) scans were preprocessed using the PreQual[36] preprocessing pipeline which performs denoising and accounts for susceptibility and eddy current artifacts and also accounts for motion. Following preprocessing, all scans underwent visual quality assurance according to the protocol described by Kim et al.[37,38] to ensure data quality prior to subsequent analyses.

### 2.2 Protocol-independent ADC representation

Under the diffusion tensor model, the diffusion-weighted signal at each voxel follows the Stejskal–Tanner relation,

$$S_i = S_0 \exp(-b_i g^T D g) \tag{1}$$

Where $S_i$ is the diffusion-weighted signal acquired with the diffusion weighting $b_i$, $S_0$ is the signal measured without any diffusion weighting, $g$ is the unit vector of the applied diffusion gradient. The term $D$ is the diffusion tensor, which is a $3 \times 3$ symmetric matrix describing the Gaussian displacement covariance of water molecules within the voxel, so that the quadratic form $g^T D g$ gives the apparent diffusivity along $g$. Rearranging Eq.1 isolates that directional diffusivity as the apparent diffusion coefficient (*ADC*),

$$\mathrm{ADC}(g) = -\frac{1}{b} \ln\left(\frac{S}{S_0}\right) = g^\top D\, g \tag{2}$$

Table 1. Acquisition parameters for the training and testing cohorts. Angular resolution spans 21 to 192 gradient directions and radial resolution spans 500 to 1500 s/mm², with both single- and multi-shell schemes represented in each split. The five testing studies were excluded entirely from training, with no subject overlap, and a single session per subject was retained.

| | **Training data (5,105 sessions)** | | | | | **Testing data (299 sessions)** | | | | |
|---|---|---|---|---|---|---|---|---|---|---|
| | **HABS-HD** | **Indiana** | **KBASE** | **ROS/MAP/MARS** | **ABCD** | **BRAIN2** | **HCP** | **Florida ADRC** | **VUIIS-MS** | **WRAP** |
| **b-value (s/mm²)** | 1000 | 1000 | 1000 | 1000 | 500, 1000 | 800 | 1000 | 1000 | 750, 1500 | 1300 |
| **Gradient directions** | 192 | 60 | 60 | 40 | 21 | 30 | 90 | 48 | 30 | 27 |
| **Sessions** | 942 | 890 | 1,396 | 801 | 1,076 | 25 | 70 | 70 | 64 | 70 |

The ADC is a convenient target for neural networks because it absorbs both the exponential attenuation and the gain factor $S_0$ which can vary per scanner, coil and protocol dependent scaling across sites and studies. This normalization results in a quantity in physical units of $mm^2/s$ that is comparable across datasets.

Since $D$ is symmetric and positive semi-definite, it has six unique entries, $d = (D_{xx}, D_{yy}, D_{zz}, D_{xy}, D_{xz}, D_{yz})$ and can therefore be recovered from ADC measurements acquired along six linearly independent directions. Expanding Equation 2 for a direction $g = (g_x, g_y, g_z)$ gives

$$ADC(g) = (g_x{}^2, g_y{}^2, g_z{}^2, 2g_xg_y, 2g_xg_z, 2g_yg_z)d \tag{3}$$

making the relationship linear in the tensor coefficients. Stacking the six measured ADCs into $a$ and the six corresponding rows of Equation 3 into a design matrix $B \in \mathbb{R}^{6\times6}$ gives the linear system $a = Bd$. Since the directions are known and fixed, $B$ is fixed as well, allowing the tensor to be recovered by

$$d = B^{-1}a \tag{4}$$

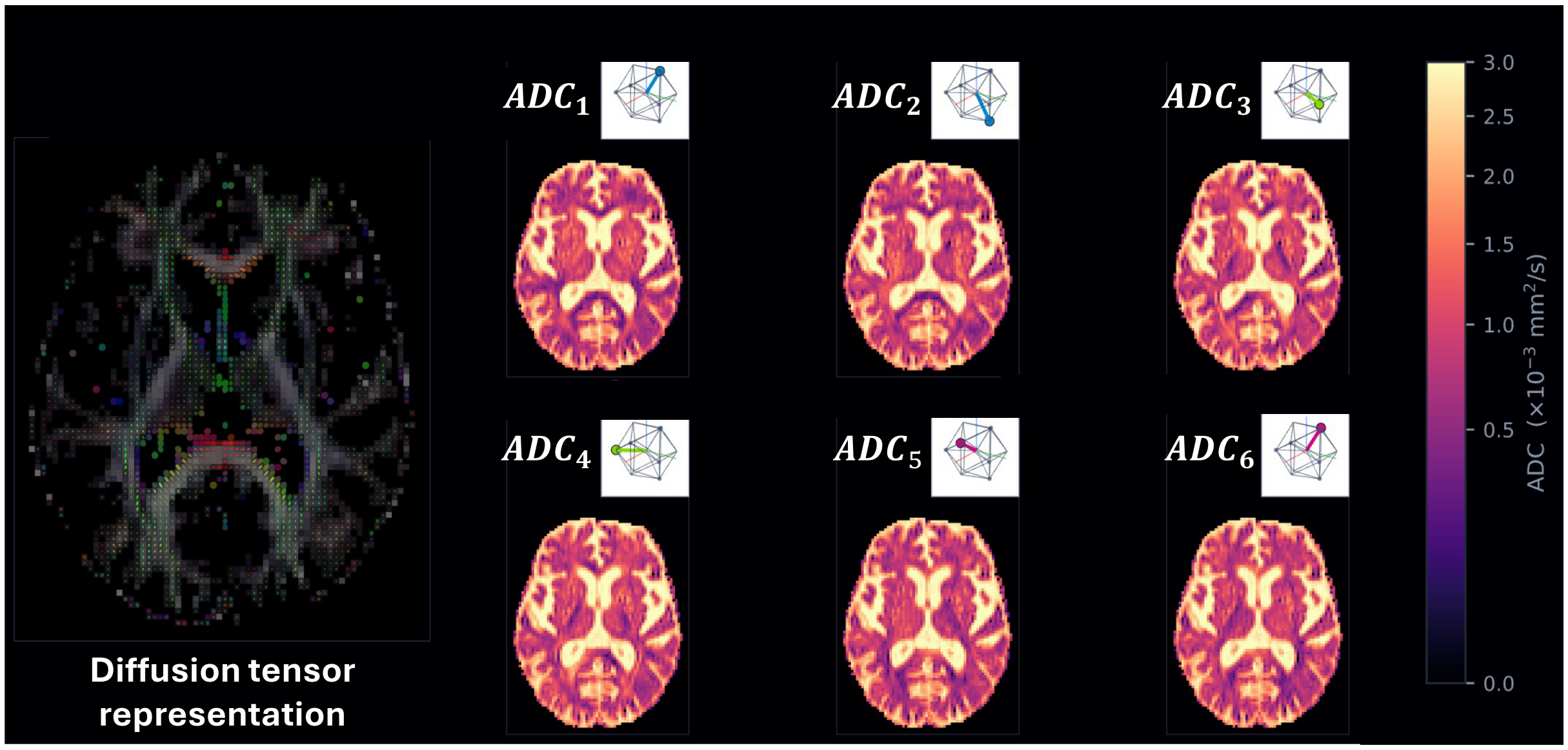

Figure 2: Because acquisitions differ in angular and radial resolution, we do not model the raw dMRI signal directly. Instead, we fit a diffusion tensor from the full dMRI (left) and project it onto six fixed icosahedral vertex directions to obtain an apparent diffusion coefficient (ADC) map per direction. Because the six directions are fixed rather than acquisition-defined, the resulting maps are anatomically consistent across protocols regardless of the original b-value or gradient scheme. The model is encouraged to learn variance–covariance structure across these six projections and at inference, the six predicted directions are enough to refit a tensor for downstream analysis.

We leverage this equivalence to construct a protocol-independent representation for dMRI FOV extension in this study. Directly learning from diffusion-weighted signals is difficult because the number of gradient directions, gradient orientations, and the applied b-values vary across acquisition protocols, resulting in inputs with varying dimensionality. We sidestep this by first fitting the diffusion tensor to each acquisition and then evaluating along six fixed directions corresponding to the six antipodal pairs of vertices of a regular icosahedron (Figure 2). These directions are maximally spread over the sphere, producing a well-conditioned design matrix B. Consequently, every acquisition, regardless of its original sampling scheme, is mapped to the same six-channel ADC representation. The network therefore learns the variance-covariance structure of these six directional ADCs rather than the acquisition-specific signal representation. The predicted ADCs are then used to reconstruct the diffusion tensor, from which standard tensor-derived measures and principal eigenvectors can be computed for downstream tractography and microstructural analysis.

### 2.3 Model inputs and conditioning

For each session, we first fit the diffusion tensor using the full acquisition and evaluate the ADC along the six canonical icosahedral directions, yielding a six-channel ADC volume $\mathbf{A} \in \mathbb{R}^{\mathbf{H}\times\mathbf{W}\times\mathbf{Z}\times\mathbf{6}}$. Prior to projection, the tensor is resampled to a common resolution of 2mm in-plane and 1mm through-plane so that all acquisitions share the same spatial resolution. The in-plane dimensions are then cropped or zero-padded to $\mathbf{128} \times \mathbf{128}$. ADC values are clipped to $\mathbf{3.5} \times \mathbf{10^{-3}}\ \mathrm{mm}^2/\mathrm{s}$ then rescaled to $[\mathbf{-1}, \mathbf{1}]$. The T1w image is percentile normalized to the 99$^{\text{th}}$ percentile before being rescaled to $[\mathbf{-1}, \mathbf{1}]$ and registered to the same grid as the diffusion data, providing slice-wise correspondence between the two modalities.

Because diffusion is a spatially continuous process, the context of the neighboring slices contains valuable contextual information for predicting the missing FOV. We therefore condition the model on the two ADC slices immediately preceding the target slice in the direction of extension, encouraging continuity in the diffusion profile along superior-inferior axis. The anatomy of the target slice is given through the co-registered T1w slice at the same slice index, allowing the model to infer the anatomy even though no diffusion measurements exist there. Let $a_z \in \mathbb{R}^{6\times128\times128}$ denote the ADC slice at index z and let $T_z \in \mathbb{R}^{1\times128\times128}$ denote the corresponding T1w slice. For an extension direction the conditioning tensor is formed by channel-wise concatenation,

$$c_z = [a_{z-2s},\ a_{z-s},\ T_z] \in \mathbb{R}^{13\times128\times128} \tag{5}$$

where $s = -1$ rcorresponds to inferior extension and $s = +1$ corresponds to the superior extension. No explicit encoding of the extension direction is provided to the network; instead, it learns to extend in either direction solely from the ordering of the conditional slices.

### 2.4 Conditional flow matching model architecture and training

Our goal is to learn the conditional distribution of the missing ADC slice given the surrounding diffusion context and the anatomy of the target slice $p_\theta(a_z \mid c_z)$. We formulate this as a conditional rectified flow problem, where the network learns a velocity field that continuously transforms Gaussian noise into the target ADC slice while remaining conditioned on $c_z$ at every step. To make the model direction agnostic, we construct training pairs symmetrically. Each volume is split at its middle slice and the slices above the midpoint are used to learn superior extension, while slices below the midpoint are used to learn inferior extension.

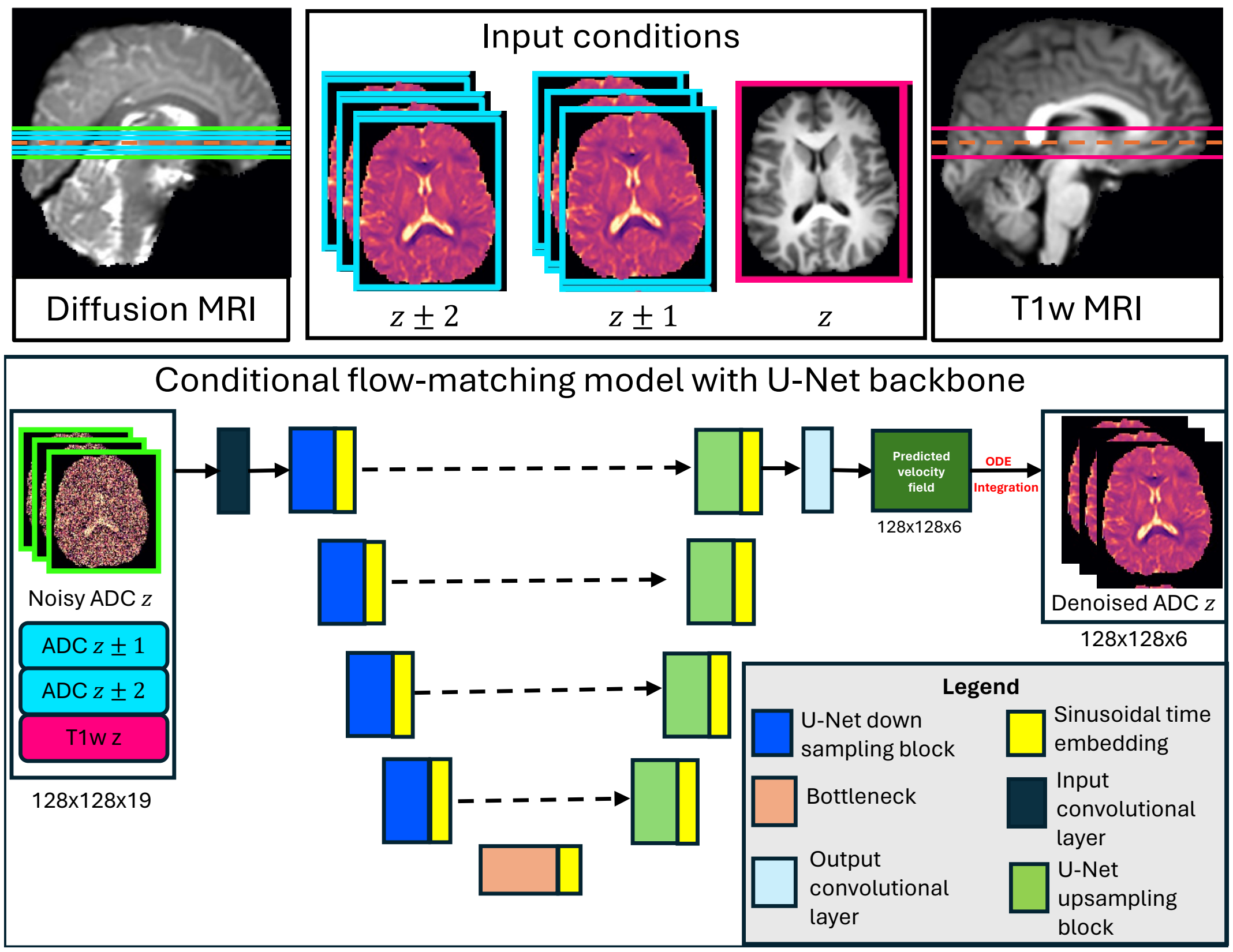

Figure 3: Top (input conditions). From the diffusion MRI, the model is conditioned on local diffusion context which are the two nearest acquired ADC slices, at offsets of 1, 2 (six directions each) and the the co-located T1w slice z from the full-FOV T1w image. The nearest ADC slices enforce diffusion continuity across the FOV boundary, while the T1w slice supplies the anatomical structure that the truncated dMRI lacks. Bottom (network). These conditions are concatenated channel-wise with the noisy ADC target and passed to a U-Net-backbone conditional flow-matching model: a 128×128×19 input and encoded through downsampling blocks to a bottleneck and decoded through upsampling blocks, with sinusoidal time embeddings injected at each stage. The network predicts a 128×128×6 velocity field, and ODE integration of that field yields the denoised six-direction ADC slice (128×128×6). The output sits in the same fixed-direction ADC space as the input, so predicted slices can be stacked back into the volume and refit to a tensor for downstream analysis.

For each training example, the clean target slice is denoted by $y_1 = a_z$, while the starting point is sampled from a standard isotropic Gaussian $\boldsymbol{y_0} = \boldsymbol{\epsilon} \sim \mathcal{N}(\boldsymbol{0}, \boldsymbol{I})$, using the standard linear interpolant, the training path becomes

$$\boldsymbol{y_t} = (1 - t)\,\boldsymbol{\epsilon} + t\,\boldsymbol{y_1}, \quad t \in [0,1] \tag{6}$$

whose velocity is constant along the trajectory,

$$v = \frac{dy_t}{dt} = y_1 - \epsilon \tag{7}$$

Because the regression target in Equation 6 is constant along the interpolation path, every point on the trajectory contributes equally during training, encouraging the network to learn nearly straight transport paths from the noise distribution to the data distribution. In practice, this allows stable generation using only a small number of integration steps during inference.

In our implementation, only the target slice is placed on the flow path, while the conditioning tensor ($c_z$) remains clean throughout training and is concatenated with $y_t$ along the channel dimension at every time step. Therefore, the network receives 19 input channels (6 noisy ADC channels and 13 conditioning channels) and predicts a 6-channel velocity field.

During training the $t \sim \mathcal{U}[0,1]$ and $\boldsymbol{\epsilon} \sim \mathcal{N}(\mathbf{0}, \boldsymbol{I})$ are sampled independently per example, and the network is optimized using the conditional flow matching objective,

$$\mathcal{L}_{\mathrm{F}} = \mathbb{E}_{\mathrm{t},\, \boldsymbol{\epsilon},\, (\boldsymbol{c_z}, \boldsymbol{y_1})}[||\boldsymbol{v}_\theta(\boldsymbol{y_t}, t, \boldsymbol{c_z}) - (\boldsymbol{y_1} - \boldsymbol{\epsilon})||_2^2] \tag{8}$$

We parameterize $v_\theta(y_t, t, c_z)$ using a four-level 2D U-Net with channel widths of doubling channel widths starting from 64 going up to 1024 corresponding to halving spatial resolutions by 2 starting from 128 going to 8 (Figure 3). The continuous time variable ($t$) is embedded using sinusoidal positional embeddings, which are projected into each resolution stage of the network. The model was trained with a batch size of 350 for 2000 epochs with a learning rate of $1e^{-4}$ on a single NVIDIA RTX 6000 Ada Generation GPU.

### 2.5 Inference and FOV extension

At inference time, the learned velocity field defined the ODE $v_\theta(y_t, t, c_z)$ which is integrated from an initial Gaussian sample, $y_0 \sim \mathcal{N}(0, I)$ at $t = 0$ to $t = 1$ using unform Euler steps,

$$y_{t+\Delta t} = y_t + \Delta t\ v_\theta(y_t, t, c_z), \qquad \Delta t = 1/N \tag{9}$$

The terminal state $\boldsymbol{y_1}$ is taken as the synthesized six-channel ADC slice. For all the results reported in this study, we use $N = 20$. FOV extension then proceeds autoregressively: each newly generated slice is appended to the ADC volume and replaces the oldest slice in the conditioning window used to predict the next slice in the extension direction. This process is repeated until the full FOV has been synthesized.

Once all missing slices have been generated, the predicted six-channel ADC representation is used to construct the diffusion tensor by solving Equation 4. The reconstructed tensor enabled computation of tensor-derived measures and principal eigenvectors used for tractography.

### 2.6 Experimental setup

We design our experiments with the goal to test the FOV extension in terms of ability to recover diffusion information that preserves downstream tractography enabled by the diffusion tensor model through white matter bundle segmentation. In addition to that, we evaluate generalizability by utilizing a separate set of dMRI datasets from acquisition protocols and studies that were entirely excluded from training. In practice, FOV truncation typically occurs in the diffusion acquisition, while the anatomical T1-weighted image remains available with full brain coverage. Therefore, during evaluation, the full-FOV T1-weighted image is provided to the model, while a portion of the diffusion volume is masked and subsequently reconstructed.

Because our method performs FOV extension autoregressively, where each generated slice becomes part of the conditioning context for subsequent predictions, errors may accumulate as the extension distance increases. To characterize this behavior, we progressively increase the missing regions in both superior and inferior directions. Specifically, we remove (10, 20, 30, 40) mm of diffusion FOV and independently bundle segmentation at each extension distance. White matter bundle segmentation evaluation is performed using TractSeg[17], which performs bundle segmentation from fiber orientation information. Because TractSeg operates on fiber orientation peaks rather than directly on diffusion tensors, we derive peaks from the reconstructed tensor by scaling the principal eigenvector ($v_1$) with FA. We compare bundle segmentations obtained from the complete ground-truth diffusion acquisition and from the FOV-extended reconstructions. Segmentation agreement is quantified using Dice similarity coefficients between the predicted and reference bundles. To determine whether FOV extension significantly affects tractography performance, we compare Dice scores obtained without FOV extension against those obtained after reconstruction using a paired t-test.

## 3. RESULTS

We report segmentation performance for the FOV extension task on seven representative TractSeg bundles spanning three anatomical regimes: two bundles confined to the inferior brain (inferior cerebellar peduncle, ICP; middle cerebellar peduncle, MCP), three traversing the full superior–inferior extent (corticospinal tract, CST; fronto-pontine tract, FPT; parieto-occipital pontine tract, POPT), and two confined to the superior brain (superior longitudinal fascicle I and II, SLF I and SLF II). For bundles localized to one end of the brain, truncation and extension were evaluated only in the corresponding direction, whereas full-extent bundles were evaluated under both superior and inferior truncation. Dice was computed against the segmentation obtained from the untruncated acquisition.

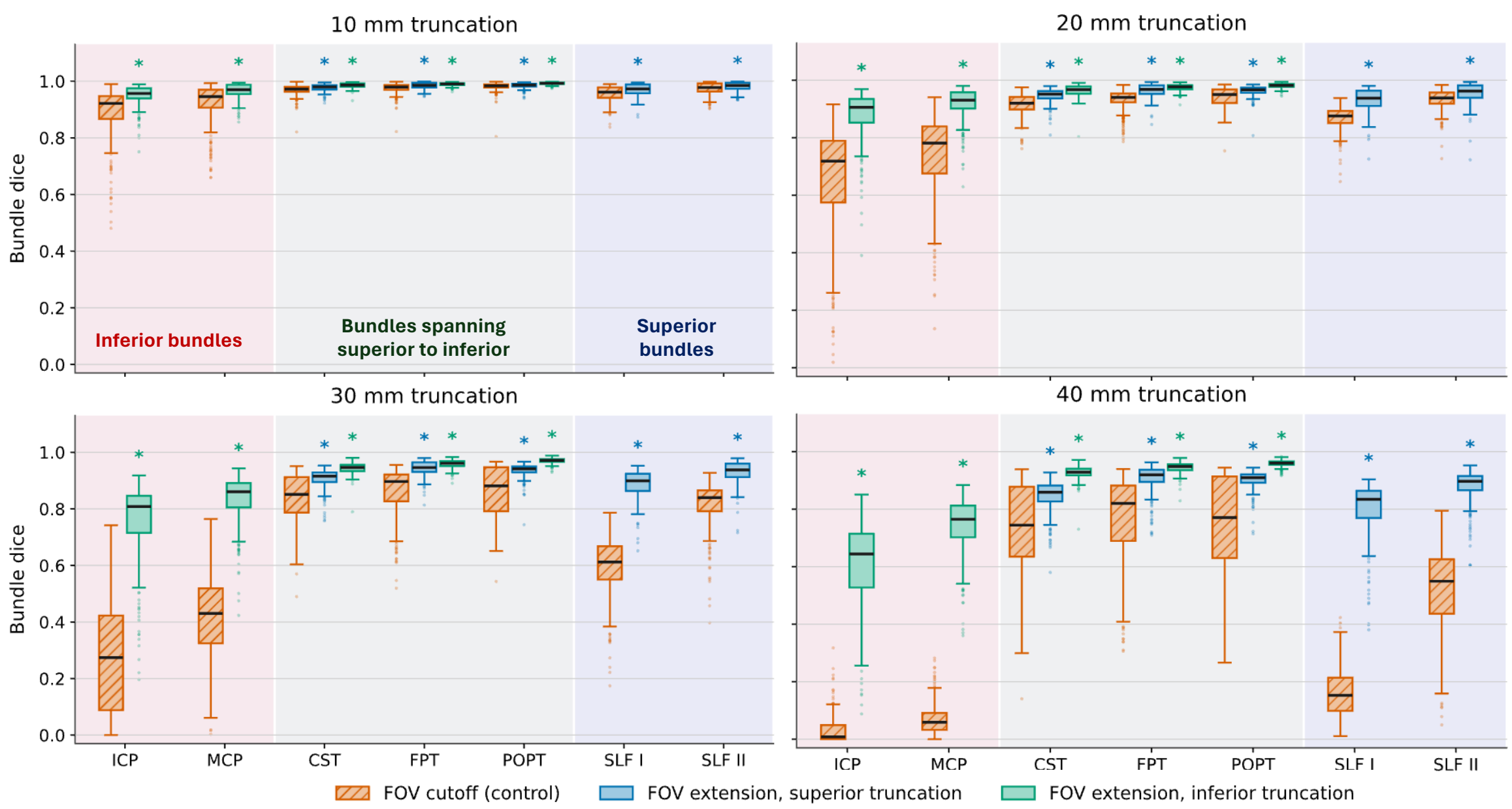

Figure 4: The benefit of FOV extension for bundle recovery grows with truncation extent, though results at the most severe truncation should be interpreted with caution. Dice agreement with the untruncated ground truth degrades for all bundles as truncation increases, but it does so far more steeply for the cutoff control than for FOV extension. At 20 and 30 mm, extension maintains median Dice above 0.8 for every bundle, whereas the control falls to 0.28 (ICP) and 0.43 (MCP) at 30 mm. At 40 mm, extension still exceeds 0.8 for most bundles, but ICP and MCP as they sit deepest in the inferior brainstem, furthest from any retained image context and are recovered only partially, with medians near 0.64 and 0.76 and wide inter-subject spread.

FOV extension significantly improved Dice over the cutoff control for every bundle and every truncation extent evaluated after correcting for multiple comparisons ($p < 0.01$), and the magnitude of the benefit grows with truncation severity (Figure 4). At 10 mm, little anatomy is removed and the control already achieves median Dice above 0.9 for all bundles, so the gains, while significant, are small. By 20 mm, truncation visibly degrades the control with ICP and MCP median dice falling to 0.72 and 0.78 respectively, while extension holds every bundle at or above 0.9. At 30 mm the control deteriorates substantially, with extension recovering a median above 0.8 for every evaluated bundle. At the most severe truncation (40 mm), several bundles are frequently not segmented at all in the control: ICP has a median Dice near zero, and MCP and SLF I fall below 0.2. Even at this truncation level, the proposed method recovers usable segmentations, with median Dice above 0.8 for all bundles except ICP and MCP, which lie deepest within the inferior brainstem and are therefore furthest from the retained image context. Figure 5 shows the same trend qualitatively with bundles that are progressively clipped at the truncation plane under FOV cutoff are restored to close agreement with the ground truth after extension, with residual disagreement concentrated at the far edge of the synthesized region.

For bundles spanning the full superior–inferior extent, inferior truncation is consistently easier to recover than superior truncation. This is because near the brainstem these pathways converge into compact, largely collinear trajectories that are well constrained by the retained slices, whereas superiorly they fan out toward the cortex and exhibit substantially greater inter-subject variability. Nevertheless, the model successfully reconstructs the bundles in both directions, and the performance gap between superior and inferior extension remains modest relative to the improvement over the cutoff control.

## 4. DISCUSSION AND CONCLUSION

In this work, we proposed and characterized a generalizable dMRI model for FOV extension. Rather than learning directly from diffusion-weighted signals whose dimensionality changes with the number of gradient directions, gradient orientations, and diffusion weightings, we first fit the diffusion tensor and project it onto six fixed icosahedral directions. This maps every acquisition onto the same six-channel ADC representation, allowing a single conditional flow model to operate across diverse protocols without retraining. We evaluated this protocol-independent representation on five independent test cohorts acquired under protocols entirely absent from training and found that the recovered FOV consistently preserved the diffusion information needed for downstream white matter analysis.

Since the motivation for FOV extension is to enable downstream analyses rather than image completion itself, we focused our evaluation on tensor-derived microstructure and tractography instead of conventional image similarity metrics. Across increasing truncation levels, the reconstructed diffusion tensors remained sufficiently faithful to recover major white matter bundles, even when the truncated acquisition alone frequently failed. As expected for an autoregressive model, performance gradually degrades as the missing FOV increases because each synthesized slice becomes the context for the next. This effect is most apparent for bundles extending deep into the missing region and for superior FOV extension, where white matter anatomy becomes increasingly dispersed and variable. Nevertheless, even under the most severe truncations, the model reliably recovered the broad morphology of the underlying white matter pathways.

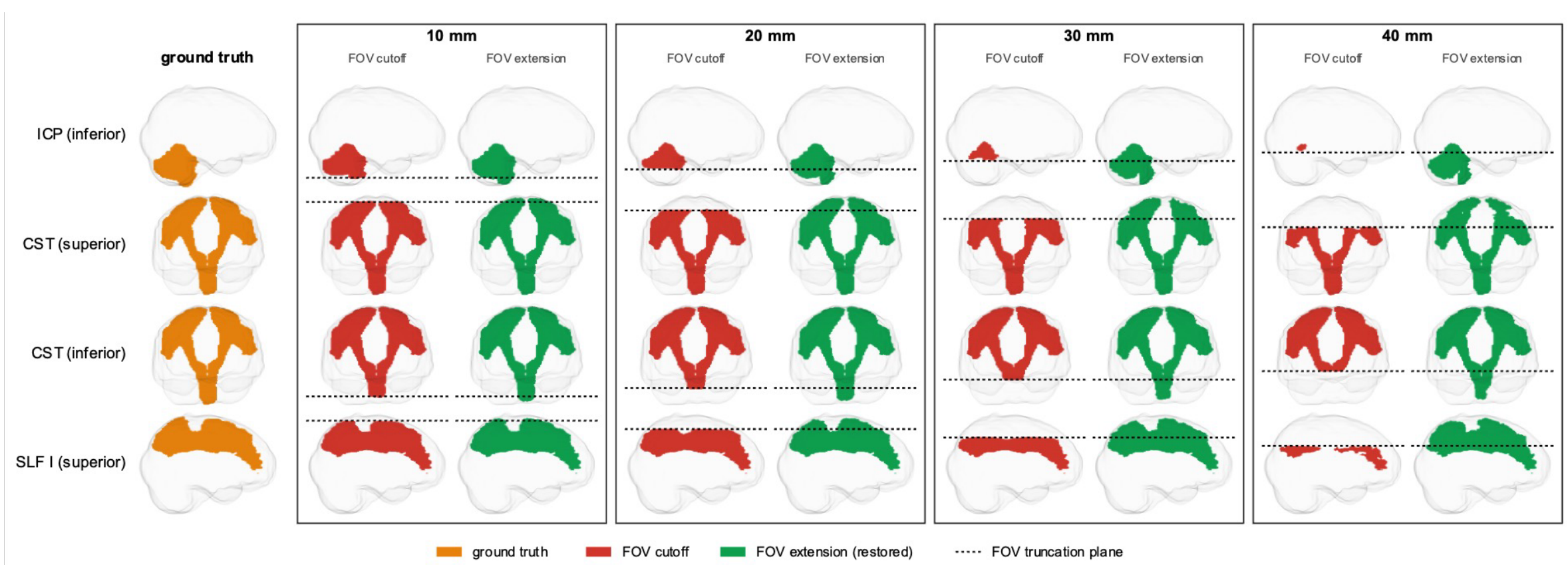

Figure 5: FOV

extension preserves the overall morphology of major white matter bundles across a range of truncation extents while maintaining subject-specific anatomical trajectories. For both superior and inferior bundles, recovery is strongest at lower truncation levels (10–30 mm), where sufficient bundle context remains in the retained slices to guide autoregressive prediction. At the most severe truncation (40 mm), where little or no local bundle context is available, the model still recovers the broad bundle morphology, although autoregressive errors accumulate with increasing distance from the observed diffusion data. This effect is particularly evident for the corticospinal tract (CST), which spans the superior–inferior extent of the brain. Although the superior and inferior FOVs are extended by the same distance, superior extension is more challenging because the white matter fans out into a broader and more variable anatomy. In contrast, inferior extension progresses into the more compact and anatomically constrained brainstem, where the retained diffusion context provides stronger guidance for autoregressive prediction, resulting in more consistent reconstructions.

The proposed framework also inherits the limitations of the diffusion tensor model used to construct the protocol-independent representation. The diffusion tensor assumes Gaussian diffusion and is therefore most appropriate for acquisitions with moderate diffusion weighting (b $\leq 1500\ s/mm^2$), where tensor fitting is well posed. Within this regime, the tensor provides a compact six-parameter representation that can be estimated from as few as six non-collinear diffusion measurements, making it broadly applicable across clinical and research datasets. Outside this regime, however, the representation becomes inadequate. Acquisitions collected exclusively at high b-values emphasize non-Gaussian diffusion that cannot be captured by a second-order tensor, while regions containing complex fiber configurations, such as crossing, fanning, or branching fibers, cannot be faithfully represented by a single dominant orientation. Consequently, the synthesized FOV is limited by the fidelity of the underlying tensor model rather than the conditional flow framework itself. Extending the same framework to fixed-dimensional representations derived from higher-order models, such as spherical harmonics or fiber orientation distributions (FODs), would remove these constraints while preserving protocol independence in future work. Additionally, future work should also validate the extension at the level of downstream analysis: whether microstructural measures recovered in the extended field of view preserve the expected associations with disease status and cognitive outcomes.

## 5. ACKNOWLEDGEMENTS

This work was supported by NIH grant R01EB017230 (PI: Landman) and the data collection and processin g for this work was supported by Alzheimer's Disease Sequencing Project Phenotype Harmonization Consortium (ADSP-PHC) that is funded by NIA (U24 AG074855, U01 AG068057 and R01 AG059716). This work was also supported by the National Cancer Institute (NCI) grants R01 CA253923 and R01 CA275015.

We gratefully acknowledge the efforts of the HABS-HD MPIs: Sid E. O'Bryant, Kristine Yaffe, Arthur Toga, Robert Rissman, and Leigh Johnson, as well as the HABS-HD Investigators: Meredith Braskie, Kevin King, James R. Hall, Melissa Petersen, Raymond Palmer, Robert Barber, Yonggang Shi, Fan Zhang, Rajesh Nandy, Roderick McColl, David Mason, Bradley Christian, Nicole Phillips, Stephanie Large, Joe Lee, Badri Vardarajan, Monica Rivera Mindt, Amrita Cheema, Lisa Barnes, Mark Mapstone, Annie Cohen, Amy Kind, Ozioma Okonkwo, Raul Vintimilla, Zhengyang Zhou, Michael Donohue, Rema Raman, Matthew Borzage, Michelle Mielke, Beau Ances, Ganesh Babulal, Jorge Llibre-Guerra, Carl Hill, and Rocky Vig. Research related to HABS-HD was supported by the National Institute on Aging of the National Institutes of Health under award numbers R01AG054073, R01AG058533, R01AG070862, P41EB015922, and U19AG078109.

Data collection and sharing for this project was provided by the Human Connectome Project (HCP; Principal Investigators: Bruce Rosen, M.D., Ph.D., Arthur W. Toga, Ph.D., Van J. Weeden, MD). HCP funding was provided by the National Institute of Dental and Craniofacial Research (NIDCR), the National Institute of Mental Health (NIMH), and the National Institute of Neurological Disorders and Stroke (NINDS). HCP data are disseminated by the Laboratory of NeuroImaging at the University of Southern California.

Data contributed from the ROS/MAP/MARS studies were supported by grants from the National Institute on Aging: R01AG017917, P30AG10161, P30AG072975, R01AG022018, R01AG056405, UH2NS100599, UH3NS100599, R01AG064233, R01AG15819, and R01AG067482, along with support from the Illinois Department of Public Health (Alzheimer's Disease Research Fund). These data are available at www.radc.rush.edu. The data were also contributed from the Wisconsin Registry for Alzheimer's Prevention was supportedby NIA AG021155, AG0271761, AG037639, and AG054047

During the preparation of this work, the authors used GitHub Copilot and ChatGPT to assist with code generation, debugging, editing, and sentence restructuring. All AI-generated content was reviewed and edited by the authors, who take full responsibility for the final content of the publication.